\documentclass[sigconf, nonacm]{acmart}
\acmConference[ICCAD '26]{IEEE/ACM International Conference on Computer-Aided Design}{November 2--6, 2026}{San Jose, CA, USA}
\acmBooktitle{Proceedings of the IEEE/ACM International Conference on Computer-Aided Design (ICCAD '26), November 2--6, 2026, San Jose, CA, USA}
\acmYear{2026}
\copyrightyear{2026}
\AtBeginDocument{}
\usepackage[english]{babel}
\usepackage{amsmath}
\usepackage[font=footnotesize]{caption}
\usepackage{graphicx}
\usepackage{float}
\usepackage{stfloats}
\usepackage{placeins}
\usepackage{xcolor}
\usepackage{multirow}
\usepackage{url}

\usepackage{booktabs}
\usepackage{makecell}
\usepackage{enumitem}
\usepackage{adjustbox}

\usepackage{pifont}

\usepackage{tikz}
\usetikzlibrary{arrows.meta, positioning, fit, calc, backgrounds, shapes.geometric, shapes.misc, shapes.symbols}

\usepackage{pgfplots}
\pgfplotsset{compat=1.18}
\usepgfplotslibrary{groupplots}
\usepackage{pgfplotstable}

\usepackage{siunitx}
\hypersetup{
  colorlinks=true,
  citecolor=blue,
  linkcolor=blue,
  urlcolor=blue,
}

\usepackage{orcidlink}
\newcommand{\IEEEauthorblockN}[1]{#1}
\newcommand{\IEEEmembership}[1]{\textit{#1}}

\begin{document}

\title[ADVERSARIAL]%
{ADVERSARIAL: And-Inverter Graph-Assisted Hardware Trojan Detection At Scale}
\titlenote{Accepted for publication at the IEEE/ACM International Conference on Computer-Aided Design (ICCAD 2026).}
\titlenote{This work was supported in part by the National Science Foundation under Grant No.~2238976, titled CAREER: Unified Reference-Free Early Detection of Hardware Trojans via Knowledge Graph Embeddings. \\
The authors are with the Department of Electrical and Computer Engineering, University of Illinois Chicago, Chicago, IL 60607 USA (e-mail: ypopry2@uic.edu, dpal2@uic.edu, vaisband@uic.edu).}

\author{\IEEEauthorblockN{%
Yaroslav Popryho\,\orcidlink{0009-0001-3382-882X},~\IEEEmembership{Graduate Student Member,~IEEE},
Debjit Pal\,\orcidlink{0000-0003-3722-5126},~\IEEEmembership{Member,~IEEE}, and
Inna Partin-Vaisband\,\orcidlink{0000-0002-6399-6672},~\IEEEmembership{Senior Member,~IEEE}}}
\affiliation{%
  \institution{}
  \city{}
  \country{}
}

\begin{abstract}
Modern System-on-Chip (SoCs) often contain hundreds of millions to tens of billions of gates, making existing Hardware Trojan (HT) detection methods impractical due to their immense scale.
The proposed approach incorporates symbolically enabled learning by modeling flattened gate-level netlists as Boolean networks represented as And-Inverter Graphs (AIGs), where all internal nodes are 2-input AND gates and inversions reside on the edges.
Each directed connection is expressed as a triple within a Knowledge Graph Embedding (KGE) framework, producing compact, constant-size per-node representations that retain multi-hop structural context.
The AIG's bounded fan-in and uniform semantics ensure training and inference complexity scale linearly with edge count, addressing major scalability bottlenecks in HT detection.
Symbolically enabled learning across deep datapaths enables the model to differentiate circuit structures from rare and functionally inconsistent connections that signify potential Trojan triggers and payloads. Experiments on large-scale SoC benchmarks demonstrate clear geometric separation between Trojan and benign nodes and practical scalability.

\end{abstract}




\maketitle
\pagestyle{plain}  
\thispagestyle{empty}

\section{Introduction}
Hardware Trojans (HTs) are malicious modifications that activate under rare conditions, changing system behavior. Typically, an HT is very small: a compact \emph{control} fragment (the trigger) that, when active, perturbs downstream \emph{datapath} logic (the payload). Locating such patterns inside modern System-on-Chip (SoCs) is extremely difficult for several reasons:

\smallskip

\noindent \textbf{Scale and design variety.}
Industrial netlists for modern SoCs often comprise hundreds of millions to several billion gates across many IP blocks~\cite {synopsys2025verification,semiwiki2022billiongate,semiengineering2019complexsocs}. As designs become progressively complex, design space grows, causing both computational complexity and space (memory) complexity of HT detection to increase exponentially, 
a long-standing obstacle noted in work on model checking, SAT, and decision diagrams~\cite{clarke2011stateexplosion,beyer2003decisiondiagrams,mcmillan2003interpolation}.  Even with optimized methods, exact reasoning over vast regions of a large netlist remains costly, and industrial flows routinely rely on abstraction, decomposition, and hardware-assisted verification to cope with state- and search-space explosion~\cite{balyo2016satrace,dao2018satfault,laeufer2024rtlautomated,semiwiki2022billiongate,synopsys2025verification}. Models that maintain a fixed-size state or rely on message passing further lose information about distant context as hop count grows---a concern when a trigger sits deep inside arithmetic logic and is separated from its payload by many levels of combinational depth~\cite{NIPS2017_f708f064,kirkpatrick2017overcomingcatastrophicforgetting}.
\smallskip

\noindent \textbf{Triggers are rare by design.}
Modern HTs often rely on signals with low activity or on deep input sequences that rarely occur during regular operation. Identifying a trigger that almost never turns on is a primary challenge. Simulation, emulation, and test-generation flows may run for a significant time and still miss those corner cases. Recent \emph{rare-event activation} efforts—including clique-based trigger activation, ATPG-guided schemes, and reinforcement learning—target this gap, yet report substantial overheads to cover large designs~\cite{lyu2020tarmac,jayasena2023atpg,pan2021tgrl}. When activation is unlikely, side-channel and optical techniques can still reveal abnormal switching or structural perturbations, but they require silicon or detailed physical design data and come with strong assumptions and non-trivial measurement costs~\cite{zhou2021optical}.

\noindent \textbf{Structure-only signals are noisy.}
Static analysis, which searches for unusual design signatures, e.g., \emph{fan-in} or \emph{fan-out},
or for low testability, is attractive because it is a fast method. However, such signatures 
frequently flag legitimate design patterns. For example, clock gating or test wrappers may have very low toggle rates or skewed fan-in/fan-out distributions, and large encoders/decoders naturally produce high fan-out. As a result, the candidate set can grow large. Reference-free screening based on controllability/observability highlights both the promise and the limits of these heuristics~\cite{salmani2017cotd}. Learning-based detectors at RTL or gate level add graph features and improve coverage, but often lack training data and are practically not generalizable~\cite{yasaei2021gnn4tj}.

These issues motivate an approach that maintains a precise yet compact representation of the logic without resorting to computationally expensive global proofs or 
message passing. The representation used in this work is the \textit{And-Inverter Graph} (AIG): every internal node is a 2-input AND gate, and each edge contains an inversion marker. This choice minimizes per-edge metadata and aligns with well-established synthesis and verification flows~\cite{brayton2010abc, biere2007aiger, biere2011aiger19}. Building on the AIG structure, each gate and connection is represented as an entity and a relation in a directed knowledge graph. This allows the netlist to be analyzed using \emph{knowledge-graph embeddings} (KGE),
which embed gates and connections into a vector space where recurring structural
patterns cluster together and unusual or inconsistent patterns stand out as outliers,
while preserving the compactness of the underlying AIG. Each directed connection is encoded as a triple (source node, relation, destination node), where the relation specifies which input pin (0 or 1) the edge corresponds to and whether the signal is inverted. A KGE model assigns a data-driven \emph{likelihood score} to each triple (and to paths) ~\cite{bordes2013transe,sun2019rotate,dettmers2018convolutional,chen2021hitter}. Repeated arithmetic and bitwise blocks form consistent local neighborhoods that the model learns efficiently. In contrast, low-activity control signals that connect into arithmetic logic in structurally atypical ways (aka \emph{control bridges}) produce unusual graph neighborhoods and receive higher anomaly scores.

Recent learning-based HT detectors at RTL and gate level (most notably GNN-based
approaches and representation-learning toolkits~\cite{yasaei2021gnn4tj}) operate
directly on circuit graphs. In contrast, the proposed framework builds
on a more structured representation. {\em First}, the AIG is interpreted as a
typed relational graph in which each edge carries explicit semantic information
(input position and polarity), enabling a far more precise description of local
logic structure. {\em Second}, the use of KGE scoring provides a lightweight
mechanism for evaluating edges and multi-hop paths without relying on costly
long-range message passing~\cite{bordes2013transe,sun2019rotate,
dettmers2018convolutional,chen2021hitter}, allowing the model to scale
naturally to very large SoCs.

\noindent{ Key contributions of the proposed framework are as follows.}

\begin{itemize}[leftmargin=10pt,labelindent=1pt,labelsep=0.5em,itemsep=2pt,topsep=2pt]
\item \textit{Scalable Analysis.}  
A near-linear-time workflow for HT detection is proposed, utilizing a uniform computational structure that efficiently maps to GPUs, enabling the processing of very large designs even under tight memory constraints.
\item \textit{AIG Normalization.}  
Gate-level netlists are transformed into compact AIGs with a small, structured set of edge types. Per-edge and multi-hop path scores are assigned by a lightweight KGE model using constant-size embeddings, which capture local logic context without requiring large Boolean states or long-range message passing.
\item {\textit{Demonstration across Trust-Hub and industrial-scale netlists.} The framework is evaluated on the public Trust-Hub TRIT-TC (combinational) and TRIT-TS (sequential) suites as well as eleven industrial SoC-style designs ranging from 1.5K to 9.87\,M nodes, and is the only HT-detection approach in this study that converges reliably on all multi-million-node netlists under a common experimental setup.} To the best of authors' knowledge, this is the first HT-detection framework demonstrated on multi-million-gate SoC designs.

\end{itemize}
  
\section{Background}
\label{sec:background}

HT detection methods that focus on side-channel analysis offer complementary visibility and have recently improved explainability through feature attribution; however, they often require high-quality power or EM data or detailed physical models. Sensitivity to noise and process variation also remains a challenge (especially considering that Trojan activation is a very rare event), and pre-silicon localization capability is inherently constrained~\cite{zhou2021optical}. Learning-based ensemble approaches can improve detection accuracy by combining multiple weak classifiers~\cite{pan_aspdac23}, yet they still depend on feature engineering and labeled data.
Crucially, side-channel methods target \emph{manufacturing-stage} insertions and can detect certain Trojans that structural methods cannot (e.g., process-variation-based attacks); conversely, the proposed structural approach operates \emph{pre-silicon} and can flag design-stage insertions before fabrication, making the two paradigms complementary rather than competing. 

Formal verification has also been explored as a golden-free alternative, but this is feasible only when the formal specification fully captures all allowed behaviors—a condition that holds only for restricted design classes~\cite{anton_date24}. Consequently, these methods require substantial specification effort, provide guarantees only within narrow domains, and cannot establish global Trojan absence. Lightweight runtime-monitoring approaches can detect Trojans during execution using analytical models~\cite{amornpaisannon_tcad24}, but they require deployed silicon and cannot flag threats pre-fabrication.

Recent work shows that HT detection is most effective when a circuit is modeled explicitly as a graph—at either RTL or gate level—so that structural irregularities can be exposed without a golden reference. Graph neural networks (GNNs)~\cite{yasaei_tcad22,yasaei2021gnn4tj} and KGE techniques~\cite{utyamishev_iscas22,utyamishev_tcad24,popryho_tcad25} exploit these representations to learn the connectivity patterns that characterize typical designs and to identify nets or subgraphs whose structure diverges from those patterns.

Building on these insights, the proposed framework adopts a similar structural viewpoint but integrates it into a unified representation that shows reliable localization of Trojan-related inconsistencies in the netlist.

\paragraph{Positioning relative to prior structural detectors.}
Several learning-based HT detectors operate on circuit graphs but differ substantially in representation, supervision, and scalability.
\emph{GNN4TJ}~\cite{yasaei2021gnn4tj} applies supervised GNN classification at RTL, requiring labeled Trojan/benign training data and does not scale to large gate-level netlists.
\emph{HW2VEC}~\cite{yu2021hw2vec} provides a general-purpose graph-learning toolkit for hardware security; its embeddings are generic and not tailored to AIG structure or Boolean semantics.
\emph{COTD}~\cite{salmani2017cotd} uses controllability/observability heuristics as structural features; it requires no learning but produces high false-positive rates on designs with legitimate low-controllability logic (e.g., clock gating, DFT).
Another approach, \emph{NetVGE}~\cite{popryho_tcad25}, operates at RTL level; ADVERSARIAL advances this to gate-level AIGs with canonical normalization, richer relation types, and robust rank aggregation.
Running HW2VEC or GNN4TJ on the same benchmarks is infeasible because HW2VEC requires RTL-level inputs and GNN4TJ requires labeled training sets; the GCN baseline operating on identical AIG-derived adjacency provides the closest matched-constraint comparison (Table~\ref{tab:detection}).

The remainder of this section summarizes the key ideas underlying the prior approaches on which the proposed framework is constructed.

\subsection{Threat model.}
\emph{Attacker.}
A malicious third-party IP vendor or outsourced design house inserts gate-level logic Trojans during design or synthesis.
The attacker may add combinational triggers (e.g., low-activity AND/OR trees gating a rarely-true condition), sequential triggers (counter- or FSM-based), or active payload logic that taps into or modifies existing datapath signals.
The attacker does \emph{not} alter the functional specification, the physical layout, or transistor-level parameters (i.e., parametric and analog Trojans are out of scope).

\smallskip

\noindent \emph{Defender.}
The defender receives the flattened gate-level netlist---standard in fabless semiconductor flows where IP is delivered as a synthesized netlist for integration and verification.
The defender has \emph{no} golden reference, \emph{no} simulation traces, and \emph{no} side-channel measurements.
Under a bounded inspection budget (top $k$\% of gates), the goal is to rank gates so that true Trojan nodes appear as early as possible in the ranked list.

\smallskip

\noindent \emph{Scope.}
The method targets Trojans whose trigger logic introduces structurally rare connectivity patterns in the AIG representation.
While a perfectly structurally isomorphic Trojan could theoretically blend in, achieving such isomorphism while successfully tapping rare payload datapath signals is practically prohibitive in dense SoCs.
The proposed method is \emph{unsupervised}: it learns only from the structural regularities present in the design under test and flags nodes whose local graph context deviates significantly from the dominant patterns.

\subsection{Structural Representation of Netlists}
To enable structural analysis, the gate-level netlist is converted into a directed graph in which each logic element is a \emph{node} and each signal connection is a directed \emph{edge}. Registers act as natural cut points, since any feedback in the design must \textbf{pass} through them; as a result, the combinational logic between registers forms acyclic regions that can be analyzed cleanly. Primary inputs, primary outputs, and register boundaries are annotated so that the resulting graph preserves the essential structure of the original circuit.

A normalized circuit representation is essential for the graph-based analyses
that follow. To this end, the netlist is converted into an AIG, which provides
a uniform and compact structure suited to large-scale reasoning: every internal
node is a 2-input AND gate and each incoming edge carries a single-bit polarity
marker. Let $\iota(e)\in\{0,1\}$ denote the polarity of edge $e$. For a node
$v$ with predecessors $u_0$ and $u_1$, the Boolean function computed at $v$ is
\begin{equation}
\label{eq:aig-def}
F(v)=\big(F(u_0)\oplus \iota(u_0\!\to\!v)\big)\wedge\big(F(u_1)\oplus \iota(u_1\!\to\!v)\big),
\end{equation}
where $\oplus$ and $\wedge$ denote XOR and AND, respectively. This
normalization fixes each node’s fan-in and limits per-edge information to a
single-polarity bit, enabling regular memory access patterns and scalable
traversal on very large SoCs. Working with a single, uniform gate representation
also makes it easier to contrast common datapath structure with rare or
irregular connections, which is crucial for isolating potential Trojan-related
logic.

\paragraph{Why AIG over gate-level graphs.}
Three properties make the AIG preferable to heterogeneous gate-level netlists for structural anomaly detection.
\emph{First}, the uniform 2-input AND semantics eliminate gate-type-specific relation encodings: a standard netlist may contain dozens of cell types, whereas the AIG vocabulary reduces to exactly four relation types ($r_{p,b}$).
\emph{Second}, structural hashing during AIG construction guarantees that functionally identical subcircuits collapse to a single representative, naturally compressing redundant datapath logic and amplifying the structural distinctiveness of Trojan insertions.
\emph{Third}, the bounded fan-in (exactly~2) ensures that the KGE neighborhood of every node has identical shape, enabling batch-parallel embedding without padding or masking.
Regarding AIG size: while a naive conversion without optimization can produce AIG blow-up for certain logic functions (e.g., XOR chains), the ABC \texttt{strash} pass used in this flow applies local rewriting (balance, refactor, rewrite) that keeps the AIG within $1.5$--$3{\times}$ the original gate count for practical circuits.

\paragraph{Decision-graph intuition for lightweight passes.}
Two ideas are borrowed from decision-graph methods without constructing full diagrams.

\emph{(i) Bottom-up reduction.} Canonical keys are assigned to AIG nodes from their children and inversion bits, so that structurally identical subgraphs are collapsed into a single representative; local identities ($\text{AND}(x,x){=}x$, $\text{AND}(x,\overline{x}){=}0$) further compress repeated arithmetic and bitwise patterns.

\emph{(ii) Input ordering.} Whether two cones normalize to the same structure is determined by the order in which their inputs are presented. Consistent orderings (e.g., adjacent bit-slices) cause related cones to be merged, while rare control signals entering at atypical points are left unmerged---compressing regular datapath logic while exposing irregular regions.

\emph{(iii) Typed triples for KGE.} Each directed AIG edge $(u\!\to\!v)$ is encoded as a triple $(u,r_{p,b},v)$, where $p\!\in\!\{0,1\}$ is the fan-in index of $v$ and $b\!\in\!\{0,1\}$ the edge inversion bit. Four relation types are obtained: $r_{0,0}$ and $r_{1,0}$ for non-inverted connections to the first and second fan-in, and $r_{0,1}$, $r_{1,1}$ for their inverted counterparts. 

\section{Methodology}

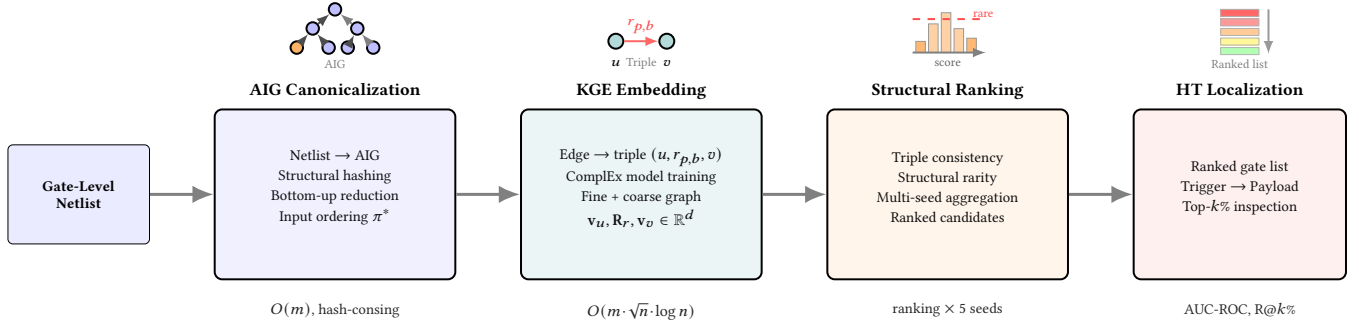
\begin{figure*}[t]
  \centering
  \resizebox{\textwidth}{!}{%
  \begin{tikzpicture}[
      >=Latex,
      font=\footnotesize,
      phase/.style={draw, rounded corners=3pt, thick, align=center,
                    minimum width=34mm, minimum height=24mm, inner sep=3pt},
      input/.style={draw, rounded corners=2pt, fill=gray!8, align=center,
                    minimum width=16mm, minimum height=7mm, font=\scriptsize},
      detail/.style={font=\scriptsize, align=center, text=black!95},
      bigarrow/.style={-{Latex[length=3mm,width=2.5mm]}, line width=1.2pt,
                       draw=black!50},
      phaselabel/.style={font=\footnotesize\bfseries, anchor=south},
  ]
  
  \node[input, fill=blue!8, minimum width=20mm, minimum height=14mm] (input) at (0,0)
      {\textbf{Gate-Level}\\\textbf{Netlist}};
  
  \node[phase, fill=blue!6, right=9mm of input] (aigphase) {};
  \node[phaselabel] at (aigphase.north) {\strut AIG Canonicalization};
  \node[detail] at ([yshift=1mm]aigphase.center) {
      Netlist $\to$ AIG\\[1pt]
      Structural hashing\\[1pt]
      Bottom-up reduction\\[1pt]
      Input ordering $\pi^*$
  };
  
  \node[phase, fill=teal!8, right=9mm of aigphase] (kgephase) {};
  \node[phaselabel] at (kgephase.north) {\strut KGE Embedding};
  \node[detail] at ([yshift=1mm]kgephase.center) {
      Edge $\to$ triple $(u,r_{p,b},v)$\\[1pt]
      ComplEx model training\\[1pt]
      Fine + coarse graph\\[1pt]
      $\mathbf{v}_u, \mathbf{R}_r, \mathbf{v}_v \in \mathbb{R}^d$
  };
  
  \node[phase, fill=orange!8, right=9mm of kgephase] (scorephase) {};
  \node[phaselabel] at (scorephase.north) {\strut Structural Ranking};
  \node[detail] at ([yshift=1mm]scorephase.center) {
      Triple consistency\\[1pt]
      Structural rarity\\[1pt]
      Multi-seed aggregation\\[1pt]
      Ranked candidates
  };
  
  \node[phase, fill=red!6, right=9mm of scorephase, minimum width=30mm] (outphase) {};
  \node[phaselabel] at (outphase.north) {\strut HT Localization};
  \node[detail] at ([yshift=1mm]outphase.center) {
      Ranked gate list\\[1pt]
      Trigger $\to$ Payload\\[1pt]
      Top-$k\%$ inspection
  };
  
  \draw[bigarrow] (input) -- (aigphase);
  \draw[bigarrow] (aigphase) -- (kgephase);
  \draw[bigarrow] (kgephase) -- (scorephase);
  \draw[bigarrow] (scorephase) -- (outphase);
  
  \node[font=\scriptsize, text=black!85, anchor=north] at ([yshift=-2mm]aigphase.south)
      {$O(m)$, hash-consing};
  \node[font=\scriptsize, text=black!85, anchor=north] at ([yshift=-2mm]kgephase.south)
      {$O(m\!\cdot\!\sqrt{n}\!\cdot\!\log n)$};
  \node[font=\scriptsize, text=black!85, anchor=north] at ([yshift=-2mm]scorephase.south)
      {ranking $\times$ 5 seeds};
  \node[font=\scriptsize, text=black!85, anchor=north] at ([yshift=-2mm]outphase.south)
      {AUC-ROC, R@$k$\%};
  
  \begin{scope}[shift={([yshift=14mm]aigphase.north)}, scale=0.45,
                every node/.style={font=\tiny}]
      \tikzstyle{nd}=[circle, draw, thick, fill=blue!25, minimum size=5pt, inner sep=0pt]
      \tikzstyle{inv}=[-latex, thick, dashed, gray]
      \tikzstyle{pedge}=[-latex, thick, black!70]
      \node[nd] (a1) at (0,0) {};
      \node[nd] (a2) at (-0.7,-0.6) {};
      \node[nd] (a3) at ( 0.7,-0.6) {};
      \node[nd, fill=orange!60] (a4) at (-1.2,-1.2) {};
      \node[nd] (a5) at (-0.2,-1.2) {};
      \node[nd] (a6) at ( 0.4,-1.2) {};
      \node[nd] (a7) at ( 1.2,-1.2) {};
      \draw[pedge] (a2)--(a1); \draw[inv] (a3)--(a1);
      \draw[pedge] (a4)--(a2); \draw[pedge] (a5)--(a2);
      \draw[pedge] (a6)--(a3); \draw[inv] (a7)--(a3);
      \node[font=\tiny, text=black!50] at (0, -1.7) {AIG};
  \end{scope}
  
  \begin{scope}[shift={([yshift=9.5mm]kgephase.north)}, scale=0.45,
                every node/.style={font=\tiny}]
      \node[circle, draw, thick, fill=teal!30, minimum size=6pt, inner sep=0pt]
          (e1) at (-0.8,0) {};
      \node[circle, draw, thick, fill=teal!30, minimum size=6pt, inner sep=0pt]
          (e2) at (0.8,0) {};
      \draw[-latex, thick, red!60] (e1) -- node[above, font=\tiny, text=red!70]{$r_{p,b}$} (e2);
      \node[font=\tiny, below=1pt of e1]{$u$};
      \node[font=\tiny, below=1pt of e2]{$v$};
      \node[font=\tiny, text=black!50] at (0,-0.7) {Triple};
  \end{scope}
  
  \begin{scope}[shift={([yshift=7.9mm]scorephase.north)}, scale=0.45,
                every node/.style={font=\tiny}]
      \foreach \i/\h/\c in {0/0.35/orange!55, 1/0.9/orange!35, 2/1.25/orange!35,
                            3/0.75/orange!35, 4/0.45/orange!55} {
          \fill[\c] (-1.0+\i*0.4, 0) rectangle (-0.7+\i*0.4, \h);
          \draw[black!40] (-1.0+\i*0.4, 0) rectangle (-0.7+\i*0.4, \h);
      }
      \draw[-latex, thick, black!50] (-1.1,0) -- (1.2,0);
      \draw[dashed, red!70, thick] (-1.1,1.05) -- (1.15,1.05);
      \node[font=\tiny, text=red!70, anchor=west] at (0.55,1.2) {rare};
      \node[font=\tiny, text=black!60] at (0, -0.35) {score};
  \end{scope}

  \begin{scope}[shift={([yshift=14mm]outphase.north)}, scale=0.45,
                every node/.style={font=\tiny}]
      \foreach \i/\c in {0/red!60, 1/red!40, 2/orange!40, 3/yellow!50, 4/green!30} {
          \fill[\c] (-0.6, -\i*0.3) rectangle (0.6, -\i*0.3-0.22);
          \draw[black!30] (-0.6, -\i*0.3) rectangle (0.6, -\i*0.3-0.22);
      }
      \node[font=\tiny, text=black!50] at (0, -1.7) {Ranked list};
      \draw[-latex, thick, black!40] (0.9, 0) -- (0.9, -1.4);
  \end{scope}
  
  \end{tikzpicture}
  }
  
  \caption{\textsc{Adversarial} pipeline overview.
  A gate-level netlist is converted into a canonical AIG via structural hashing, bottom-up reduction, and input ordering ($O(m)$).
  AIG edges are encoded as typed triples and used to train a ComplEx KGE model on both fine-grained and congruence-compressed graphs.
  Triple-level consistency and structural rarity are aggregated across multiple seeds to produce stable rankings.
  The output is a ranked list of suspect gates for bounded-budget analyst inspection.}
  \label{fig:overall-workflow}
  \end{figure*}

The proposed methodology relies on the fact that {\bf\em canonical symbolic
representations of Boolean functions preserve structural characteristics that
can be exploited for HT detection}. In particular, the construction of AIGs in
the proposed flow uses bottom-up reduction, structural sharing of congruent
subgraphs, and carefully selected input-variable orderings to obtain a compact
and stable representation. Together, these techniques provide a substrate for symbolically enabled statistical learning on which the Hitter-based KGE detection algorithm operates (Fig.~\ref{fig:overall-workflow}).

A digital circuit $C$ is represented as a directed graph $G = (V,E)$ in which gates and terminals are nodes, and wires are directed edges. Nodes include primary inputs (PI), primary outputs (PO), combinational gates (INV, NAND, NOR, XOR, etc). An edge $(u \to v)$ is created whenever the signal produced by $u$ serves as an input to $v$. In AIG-style designs, each edge is labeled with a polarity $s_{uv} \in \{+1,-1\}$ (non-inverting or inverting). This label is used by the encoder but does not change the graph topology.

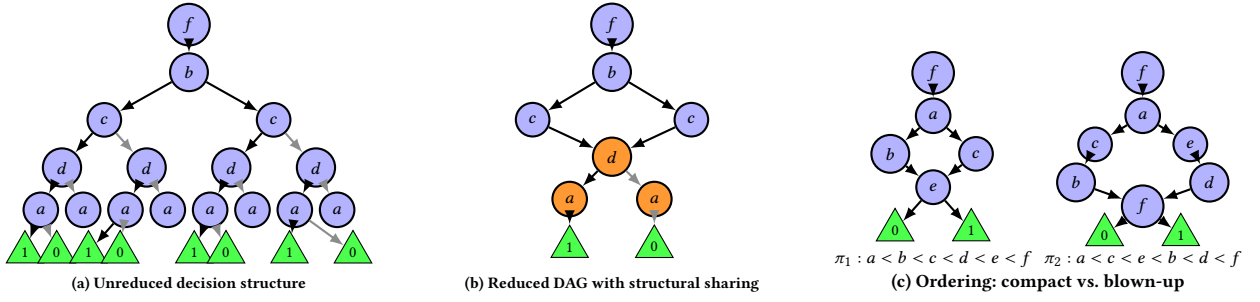
\begin{figure*}[t]
  \centering
  \vspace{-10pt}
  \begin{minipage}[t]{0.31\textwidth}
  \centering
  \begin{tikzpicture}[scale=0.7, every node/.style={font=\scriptsize}]
  \tikzstyle{var}=[circle, draw, thick, fill=blue!30, minimum size=11pt]
  \tikzstyle{term}=[regular polygon, regular polygon sides=3, draw,
                    fill=green!70, inner sep=1pt]
  \tikzstyle{edge0}=[-latex, thick, gray!90]
  \tikzstyle{edge1}=[-latex, thick, black]
  
  \node[var] (f) at (0,3.2) {$f$};
  \node[var] (b) at (0,2.3) {$b$};
  
  \node[var] (cL) at (-1.6,1.4) {$c$};
  \node[var] (cR) at ( 1.6,1.4) {$c$};
  
  \node[var] (d11) at (-2.4,0.5) {$d$};
  \node[var] (d12) at (-0.8,0.5) {$d$};
  \node[var] (d21) at ( 0.8,0.5) {$d$};
  \node[var] (d22) at ( 2.4,0.5) {$d$};
  
  \node[var] (a11) at (-2.8,-0.3) {$a$};
  \node[var] (a10) at (-2.0,-0.3) {$a$};
  \node[var] (a21) at (-1.2,-0.3) {$a$};
  \node[var] (a20) at (-0.4,-0.3) {$a$};
  \node[var] (a31) at ( 0.4,-0.3) {$a$};
  \node[var] (a30) at ( 1.2,-0.3) {$a$};
  \node[var] (a41) at ( 2.0,-0.3) {$a$};
  \node[var] (a40) at ( 2.8,-0.3) {$a$};
  
  \foreach \x/\name/\txt in {-3.1/t11/1, -2.5/t10/0, -1.9/t21/1, -1.3/t20/0,
                             0.1/t31/1, 0.7/t30/0, 1.9/t41/1, 3.1/t40/0}
  {
    \node[term] (\name) at (\x,-1.1) {\txt};
  }
  
  \draw[edge1] (f) -- (b);
  \draw[edge1] (b) -- (cL);
  \draw[edge1] (b) -- (cR);
  
  \draw[edge1] (cL) -- (d11);
  \draw[edge0] (cL) -- (d12);
  \draw[edge1] (cR) -- (d21);
  \draw[edge0] (cR) -- (d22);
  
  \foreach \d/\aone/\azero in {d11/a11/a10, d12/a21/a20,
                               d21/a31/a30, d22/a41/a40}
  {
    \draw[edge1] (\d) -- (\aone);
    \draw[edge0] (\d) -- (\azero);
  }
  
  \foreach \a/\tone/\tzero in {a11/t11/t10, a21/t21/t20,
                               a31/t31/t30, a41/t41/t40}
  {
    \draw[edge1] (\a) -- (\tone);
    \draw[edge0] (\a) -- (\tzero);
  }
  
  \node at (0,-1.7) {\textbf{(a) Unreduced decision structure}};
  \end{tikzpicture}
  \end{minipage}
  \begin{minipage}[t]{0.31\textwidth}
  \centering
  \begin{tikzpicture}[scale=0.7, every node/.style={font=\scriptsize}]
  \tikzstyle{var}=[circle, draw, thick, fill=blue!30, minimum size=11pt]
  \tikzstyle{shared}=[circle, draw, thick, fill=orange!80, minimum size=11pt]
  \tikzstyle{term}=[regular polygon, regular polygon sides=3, draw,
                    fill=green!70, inner sep=1pt]
  \tikzstyle{edge0}=[-latex, thick, gray!90]
  \tikzstyle{edge1}=[-latex, thick, black]
  
  \node[var]    (f2) at (0,3.2) {$f$};
  \node[var]    (b2) at (0,2.3) {$b$};
  
  \node[var]    (cL2) at (-1.5,1.4) {$c$};
  \node[var]    (cR2) at ( 1.5,1.4) {$c$};
  
  \node[shared] (dS)  at (0,0.7) {$d$};
  \node[shared] (aS1) at (-0.8,-0.1) {$a$};
  \node[shared] (aS0) at ( 0.8,-0.1) {$a$};
  
  \node[term] (t1) at (-0.8,-1.0) {1};
  \node[term] (t0) at ( 0.8,-1.0) {0};
  
  \draw[edge1] (f2) -- (b2);
  \draw[edge1] (b2) -- (cL2);
  \draw[edge1] (b2) -- (cR2);
  
  \draw[edge1] (cL2) -- (dS);
  \draw[edge1] (cR2) -- (dS);
  
  \draw[edge1] (dS) -- (aS1);
  \draw[edge0] (dS) -- (aS0);
  
  \draw[edge1] (aS1) -- (t1);
  \draw[edge0] (aS0) -- (t0);
  
  \node at (0,-1.7) {\textbf{(b) Reduced DAG with structural sharing}};
  \end{tikzpicture}
  \end{minipage}
  \begin{minipage}[t]{0.31\textwidth}
  \centering
  \begin{tikzpicture}[scale=0.63, every node/.style={font=\scriptsize}]
  \tikzstyle{var}=[circle, draw, thick, fill=blue!30, minimum size=10pt]
  \tikzstyle{term}=[regular polygon, regular polygon sides=3, draw,
                    fill=green!70, inner sep=1pt]
  \tikzstyle{edge}=[-latex, thick]
  
  \node[var] (fL) at (-2.2,2.4) {$f$};
  \node[var] (aL) at (-2.2,1.5) {$a$};
  \node[var] (bL) at (-3.1,0.7) {$b$};
  \node[var] (cL) at (-1.3,0.7) {$c$};
  \node[var] (eL) at (-2.2,0.0) {$e$};
  \node[term] (0L) at (-3.0,-0.9) {0};
  \node[term] (1L) at (-1.4,-0.9) {1};
  
  \draw[edge] (fL)--(aL);
  \draw[edge] (aL)--(bL);
  \draw[edge] (aL)--(cL);
  \draw[edge] (bL)--(eL);
  \draw[edge] (cL)--(eL);
  \draw[edge] (eL)--(0L);
  \draw[edge] (eL)--(1L);
  
  \node at (-2.2,-1.5) {\scriptsize $\pi_1: a<b<c<d<e<f$};
  
  \node[var] (fR) at (2.2,2.4) {$f$};
  \node[var] (aR) at (2.2,1.5) {$a$};
  
  \node[var] (cR1) at (1.2,0.9) {$c$};
  \node[var] (eR1) at (3.2,0.9) {$e$};
  \node[var] (bR1) at (0.8,0.1) {$b$};
  \node[var] (dR1) at (3.6,0.1) {$d$};
  \node[var] (fR2) at (2.2,-0.4) {$f$};
  
  \node[term] (0R) at (1.4,-1.0) {0};
  \node[term] (1R) at (3.0,-1.0) {1};
  
  \draw[edge] (fR)--(aR);
  \draw[edge] (aR)--(cR1);
  \draw[edge] (aR)--(eR1);
  \draw[edge] (cR1)--(bR1);
  \draw[edge] (eR1)--(dR1);
  \draw[edge] (bR1)--(fR2);
  \draw[edge] (dR1)--(fR2);
  \draw[edge] (fR2)--(0R);
  \draw[edge] (fR2)--(1R);
  
  \node at (2.2,-1.5) {\scriptsize $\pi_2: a<c<e<b<d<f$};
  
  \node at (0,-2.0) {\footnotesize\textbf{(c) Ordering: compact vs.\ blown-up}};
  \end{tikzpicture}
  \end{minipage}
  \vspace{-5pt}
  \caption{
  Transformations in the proposed HT-detection flow:
  (a) unreduced logic with duplicated subtrees;
  (b) bottom-up reduction with structural sharing (merged cones in orange);
  (c) impact of input ordering—$\pi_1$ yields a compact AIG, whereas $\pi_2$ causes a size blow-up.
  }
  \vspace{-5pt}
  \label{fig:single-row-symbolic}
  \end{figure*}
  
\paragraph{Bottom-up reduction.}
In the proposed flow, AIG construction implicitly 
identifies structurally identical subgraphs and represents them with a single node. Formally, two nodes $x$ and $y$ labeled with AND gates satisfy
\[
x \equiv y
\;\;\Longleftrightarrow\;\;
\{\!\{ (x_1,x_2), (x_2,x_1) \}\!\} = 
\{\!\{ (y_1,y_2), (y_2,y_1) \}\!\},
\]
where edge negations (inversion flags) are taken into account (Fig. \ref{fig:single-row-symbolic}a-b). Thus, identical
subfunctions yield identical AIG nodes. Let $\mathcal{U}$ denote the global
structural-hash table. During construction:
\[
\begin{aligned}
(g, g_1, g_2) \in \mathcal{U} &\;\Rightarrow\; \text{return existing representative},\\
(g, g_1, g_2) \notin \mathcal{U} &\;\Rightarrow\; \text{insert and continue}.
\end{aligned}
\]

\paragraph{Input ordering and representational complexity.}
The input-variable ordering $\pi$ used during AIG construction influences how
much structural sharing can be exposed by structural hashing. Let
$A_\pi(C)$ denote the AIG obtained from circuit $C$ under ordering $\pi$, and
let $\mathcal{N}_\pi$ be its node set. For a fixed $\pi$, define an equivalence
relation $u \equiv_\pi v$ over $\mathcal{N}_\pi$ such that
$u \equiv_\pi v$ when $u$ and $v$ have the same gate type and identical
multisets of fanin literals (children and polarities, up to permutation). Each
equivalence class
\[
[v]_\pi = \{ u \in \mathcal{N}_\pi : u \equiv_\pi v \}
\]
corresponds to one structurally unique subfunction represented after hashing.
The amount of reuse exposed by ordering $\pi$ can therefore be quantified as
\[
R(\pi) = \sum_{v \in \mathcal{N}_\pi}
\mathbf{1}\!\left[\,|[v]_\pi| > 1\,\right],
\]
the number of nodes belonging to nontrivial equivalence classes. A desirable
ordering maximizes $R(\pi)$, yielding a smaller AIG (as shown on Fig. \ref{fig:single-row-symbolic}):
\[
\pi^\ast = \arg\max_\pi R(\pi),
\qquad
|A_{\pi^\ast}(C)| = |\{[v]_{\pi^\ast}\}|.
\]

\paragraph{Symbolic representation for functional equivalence.}
Let $F$ and $G$ be two Boolean functions over the same variable set.  Three
approaches exist for deciding $F \equiv G$:

\begin{enumerate}
[leftmargin=15pt,labelindent=1pt,labelsep=0.5em,itemsep=2pt,topsep=2pt]
\item Exhaustive truth-table comparison: check $F(\mathbf{x}) = G(\mathbf{x})$
for all $\mathbf{x} \in \{0,1\}^n$, which is exponential in $n$.

\item Algebraic manipulation using Boolean identities (De Morgan, absorption,
Shannon expansion, etc.). In practice, the required rewrite sequence suffers
from the \emph{phase-ordering problem}, with exponentially many possible
rewrite paths and low chance of choosing a good one.

\item Symbolic representation via canonical AIG forms. If reduced graphs
$R_\pi(F)$ and $R_\pi(G)$ built under the same ordering $\pi$ satisfy
$R_\pi(F) \cong R_\pi(G)$, then $F \equiv G$ follows immediately.
\end{enumerate}

The proposed methodology adopts the third approach.  
AIGs of the original circuit $C$ and its canonicalized version $C'$  are constructed under $\pi^\ast$, yielding DAGs $A_\pi(C)$ and $A_\pi(C')$.  
Here, $C$ denotes the circuit under analysis, and $C'$ is a functionally
equivalent representation obtained by applying a controlled sequence of
canonicalizing transformations. Comparing the AIGs $A_\pi(C)$ and $A_\pi(C')$ therefore highlights exactly
those regions whose structure fails to stabilize under canonicalization. 

Let $\mathcal{M}(A_\pi(C),A_\pi(C'))$ denote the set of maximal common subgraph isomorphisms.  
This set captures all pairs of
subgraphs $(G,G')$ such that
\[
G \cong G'
\quad\Longrightarrow\quad
\llbracket G \rrbracket = \llbracket G' \rrbracket,
\]
where $\llbracket \cdot \rrbracket$ denotes the Boolean function encoded by the subgraph.
Any region where no isomorphism exists represents a functional deviation and is thus a candidate HT location.
The fan-in cone of such a node—whose canonical AIG representation under ordering $\pi^*$ differs from the corresponding cone in the original AIG—is referred to as a \emph{suspect cone}.

\paragraph{KGE formulation}
The AIG $A_\pi(C)$ is lifted into a KGE setting by mapping each AIG node to an entity $e_i$ and each edge relation (fanin index and polarity) to a relation type $r_j$.  
Let $\mathcal{G} = (E,R,T)$ denote the resulting graph, with triples
\[
T = \{ (e_i, r_j, e_k) \mid \text{corresponding AIG edge exists} \}.
\]
The embedding model is trained to discriminate observed AIG edges from absent ones: it learns vectors $\mathbf{v}_{e_i},\mathbf{v}_{e_k}\!\in\!\mathbb{R}^d$ and relation operators $\mathbf{R}_j$ so that the scoring function $f$ ranks any true triple above negative (corrupted) samples drawn by replacing one endpoint,
\[
f(e_i,r_j,e_k) \approx 1
\quad\text{iff}\quad
(e_i,r_j,e_k) \in T.
\]
Intuitively, well-supported AIG connections (those that repeat across the datapath) receive high scores, while structurally inconsistent edges—characteristic of Trojan triggers—receive low ones, providing the per-edge signal used for ranking.

HT detection is then formulated as the identification of minimal hitting sets of inconsistent triples:
\[
\mathcal{H}^\ast = 
\arg\min_{\mathcal{H}}
\left\lvert \mathcal{H} \right\rvert
\quad\text{s.t.}\quad
\forall t \in T' \setminus \mathcal{H}:\; f(t) \ge \tau,
\]
where $T'$ is the triple set derived from the suspect design $C'$ and $\tau$ is a consistency threshold.  
This mirrors the extraction of a minimal UNSAT core in SAT: inconsistent embedding constraints point to the region where structural behavior deviates from what is typically observed.

\paragraph{Weighted--attentive HittER-based scoring.}
To emphasize structurally uncommon AIG connections during embedding, the framework uses the weighted--attentive softmax introduced in NetVGE~\cite{popryho_tcad25}.
For every typed triple $(u,r,v)$ extracted from the canonical AIG, the KGE model
computes an unnormalized score
\[
s(u,r,v)=\mathbf{v}_u^{\top}\mathbf{R}_r\,\mathbf{v}_v ,
\]
where $\mathbf{v}_u,\mathbf{v}_v\!\in\!\mathbb{R}^d$ are node embeddings and
$\mathbf{R}_r$ is the learned operator for relation type $r$.

Instead of a standard softmax, HittER applies an \emph{attentive} weighting over
candidate triples:
\[
P(u,r,v)
=\frac{\exp\!\big(\alpha_{u,r,v}\,s(u,r,v)\big)}
{\sum_{(u,r,v')\in\mathcal{N}(u,r)}\exp\!\big(\alpha_{u,r,v'}\,s(u,r,v')\big)},
\]
where the attention coefficient $\alpha_{u,r,v}\!\in\![0,1]$ is a learned weight
that down-weights frequent or structurally “routine’’ patterns and up-weights
rare ones.  
As shown in~\cite{popryho_tcad25}, this weighted–attentive
normalization acts as an adaptive prior:  
\[
\begin{aligned}
\alpha_{u,r,v}\!\uparrow &\;\Rightarrow\;
    \text{rare or atypical patterns receive higher weight},\\[2pt]
\alpha_{u,r,v}\!\downarrow &\;\Rightarrow\;
    \text{routine patterns are down-weighted}.
\end{aligned}
\]

Applied to AIGs, this mechanism naturally highlights edges whose structural
context deviates from the dominant datapath patterns. Regular arithmetic and
control cones receive low attention weights (high redundancy), whereas
Trojan-like trigger connections—typically sparse, asymmetric, and weakly aligned
with surrounding logic—receive higher attention and therefore stand out during training.

\paragraph{Complexity.}
Let $n{=}|V|$ and $m{=}|E|$ denote the node and edge counts of the AIG.
(i)~AIG construction via structural hashing: $O(m)$ time and space using \emph{hash-consing} ---a canonicalization technique in which every newly constructed node is looked up in a global hash table keyed by its (children, polarity) tuple so that structurally identical subgraphs are never materialized twice. This is the same mechanism used by ABC's \texttt{strash} pass~\cite{brayton2010abc,mishchenkostandardizing}.
(ii)~Bottom-up reduction: $O(m)$ via a single reverse-topological pass.
(iii)~KGE training: each epoch processes all $m$ triples; with embedding dimension $d{=}O(\!\sqrt{n})$ and $E{=}O(\log n)$ epochs (adaptive schedule), total cost is $O(m\sqrt{n}\log n)$.
(iv)~Scoring and rank aggregation: $O(n\log n)$ for sorting-based ranking.
The overall pipeline is dominated by KGE training at $O(m\sqrt{n}\log n)$, which is near-linear for sparse AIG graphs where $m{=}2n$ (each node has fan-in~2).
\vspace{-3mm}
\paragraph{Trojan detection in the reduced latent space}
Node embeddings produced by the KGE model are projected via t-SNE—an approach
well suited for preserving local structure in high-dimensional and highly
imbalanced datasets~\cite{maaten2008visualizing}. The resulting projections
reveal distinct geometric roles for different node types. \textit{Payload}
nodes occupy the centers of dense clusters, reflecting their integration into
regular compute logic. In contrast, \textit{trigger} candidates lie near
cluster boundaries or form small off-manifold groups, reflecting their
structurally rare and weakly integrated logic patterns. This geometric
separation between cluster centers and peripheral regions provides an
interpretable basis for distinguishing trigger and payload logic in very large
SoCs.
\section{Experimental Results}

\begin{figure*}[!b]
    \centering
    \begin{minipage}[b]{0.49\textwidth}
        \centering
        \includegraphics[width=\columnwidth]{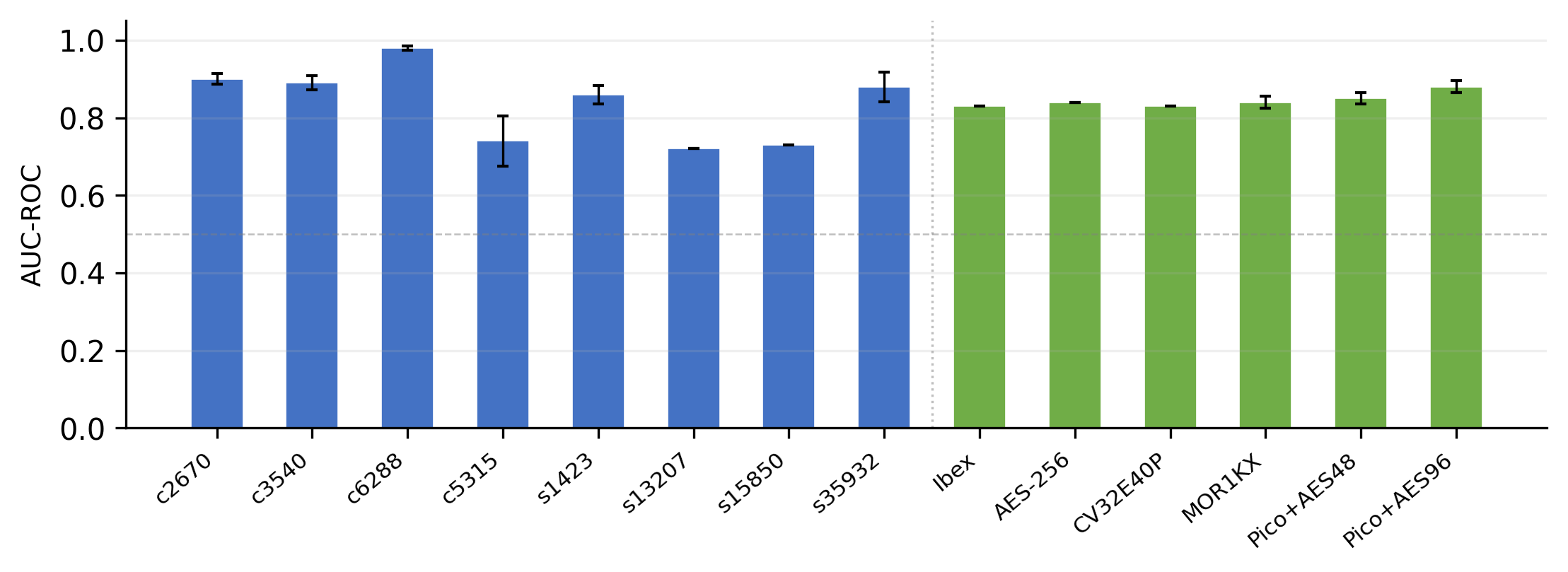}\\[0.35em]
        {\footnotesize
        \textcolor[HTML]{4472C4}{\rule{3.8pt}{3.8pt}}\,\textit{Trust-Hub}\qquad
        \textcolor[HTML]{70AD47}{\rule{3.8pt}{3.8pt}}\,\textit{Industrial SoCs}}
        \captionof{figure}{AUC-ROC per circuit family (mean $\pm$ std over five Trojan variants), including Trust-Hub and industrial SoC designs.}
        \label{fig:auc-per-family}
    \end{minipage}\hfill
    \begin{minipage}[b]{0.49\textwidth}
        \centering
        \includegraphics[width=\columnwidth]{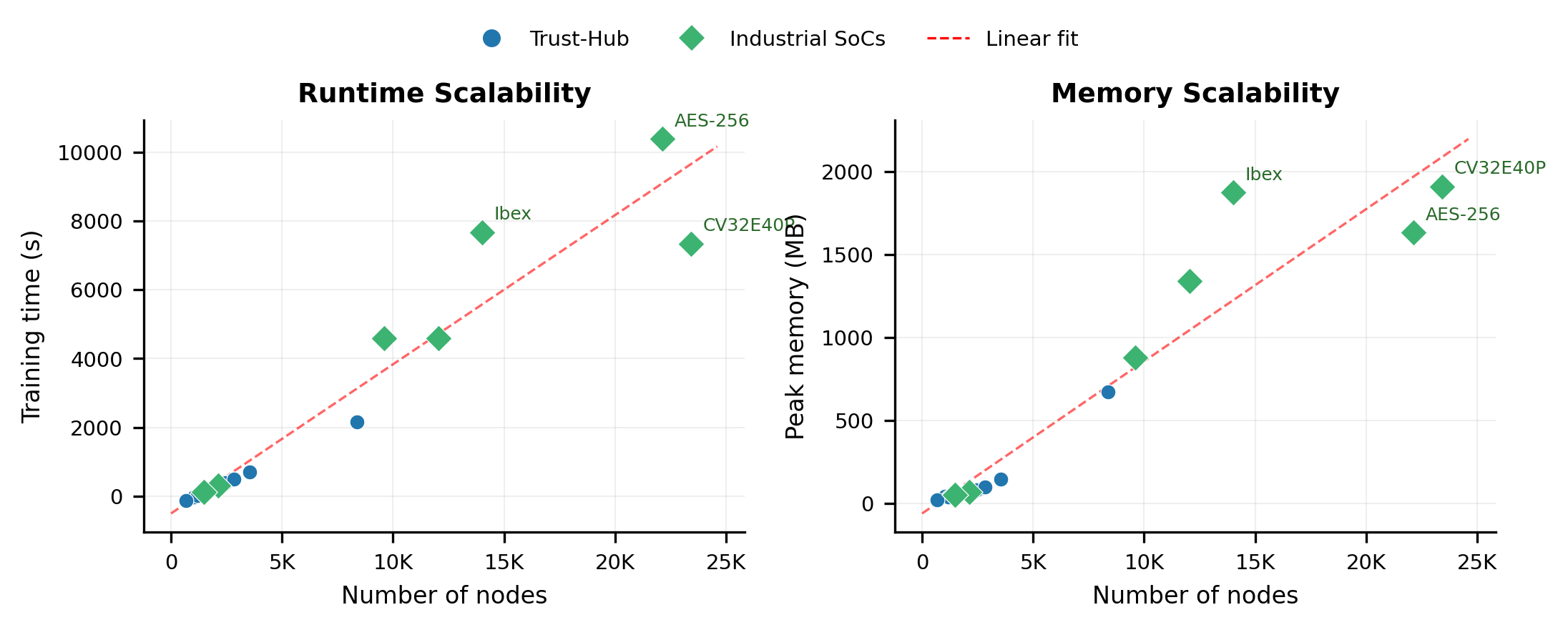}
        \captionof{figure}{Runtime and memory scalability of the \textsc{Adv.} pipeline (single CPU core). Both scale linearly with the number of nodes, confirming feasibility as a screening pass.}
        \label{fig:scalability}
    \end{minipage}
\end{figure*}

\subsection{Setup}
\label{sec:setup}

\noindent\textbf{Benchmarks.}
All publicly available Trust-Hub benchmarks (TRIT-TC and TRIT-TS)~\cite{trusthub} are used, with Trojans inserted in both datapath and control logic. Scalability is assessed on eleven industrial SoC-style netlists taken from~\cite{popryho2025automatedhardwaretrojaninsertion}, listed in Table~\ref{tab:date_sizes}; the suite spans four orders of magnitude in size, from a 1.5\,K-node UART controller to a 9.87\,M-node PicoRV32+AES96 SoC. 
{\bf\em Each design is injected with multiple trigger families} (sequential counters, Hamming-weight comparators, glitch chains, hash-based triggers) and with {\bf\em payload types} including bit-flip, signal-leak, stuck-at, and reset-disable. SoC-style designs and their trigger \& payload mechanisms are publicly released on OSF\footnote{\url{https://osf.io/y53bq/overview?view_only=75358a728c2f4a51885226a29b5ac34d}}.

\begin{table}[t]
\centering
  \caption{Industrial SoC benchmarks from~\cite{popryho2025automatedhardwaretrojaninsertion}. Designs span four orders of magnitude in size.}
  \vspace{-5pt}
\label{tab:date_sizes}
\setlength{\tabcolsep}{3pt}
\renewcommand{\arraystretch}{0.9}
\resizebox{\columnwidth}{!}{%
\begin{tabular}{@{}lrrrl@{}}
\toprule
\textbf{Design} & \textbf{Nodes} & \textbf{Edges} & \textbf{HT} & \textbf{Description} \\
\midrule
UART16550      & 1.5K   & 3.3K   & 8   & Serial UART controller \\
ETH-1G         & 2.1K   & 4.6K   & 10  & Gigabit Ethernet MAC \\
PicoRV32       & 9.6K   & 21.1K  & 18  & RISC-V CPU \\
SHA-256        & 12K    & 26.4K  & 22  & Cryptographic hash \\
Ibex           & 14K    & 32.2K  & 26  & RISC-V CPU core \\
AES-256        & 22K    & 39K    & 26  & Symmetric cipher \\
CV32E40P       & 23K    & 52.5K  & 33  & PULP RISC-V core \\
MOR1KX         & 96K    & 160K   & 93  & OpenRISC CPU \\
PicoRV32+AES48 & 2.47M  & 3.50M  & 111 & SoC: CPU + 48b AES \\
PicoRV32+AES96 & 9.87M  & 14.06M & 171 & SoC: CPU + 96b AES \\
\bottomrule
\vspace{-20pt}
\end{tabular}%
}
\end{table}

\smallskip
\noindent\textbf{AIG construction.}  RTL designs are first synthesized to flattened gate-level netlists using Synopsys DC, and the resulting netlists are then lowered to AIGs. Structural hashing, local rewriting, and input normalization are performed by ABC's \texttt{strash}~\cite{brayton2010abc,mishchenkostandardizing}. The canonical node and \texttt{edgelist} representation (each gate's two inputs, edge inversions, and topological position) is then converted into the typed triples consumed by the KGE model. The variable ordering $\pi$ is obtained by a topological traversal from primary inputs ($O(n)$); although computing the optimal $\pi^\ast$ is NP-hard, the heuristic ordering proves sufficient in practice: varying the tie-breaking seed across three restarts of the AIG construction shifts downstream detection AUC-ROC by less than $0.02$ on every benchmark, indicating that the proposed pipeline is insensitive to the specific choice of $\pi$.

\smallskip
\noindent\textbf{Baselines.}
Two matched-constraint structural baselines are compared against \textsc{Adversarial}: a \emph{KGE} model (ComplEx ~\cite{utyamishev_tcad24}, with no fusion, no adaptive capacity, and no multi-seed aggregation) and a \emph{GCN} ~\cite{yasaei_tcad22} trained on the same AIG-derived adjacency. Both baselines are fed identical inputs under identical compute budgets, so differences in ranking quality are attributable to the proposed AIG+KGE pipeline itself. Supervised RTL-level detectors (GNN4TJ, HW2VEC) are excluded by construction: labeled training data and/or RTL-level inputs are required, neither of which is permitted under the golden-free threat model (Section~\ref{sec:background}).

\smallskip
\noindent\textbf{Metrics.}
The operational goal of golden-free HT screening is to \emph{rank} gates so that a tiny fraction of the netlist can be inspected while most Trojan logic is still encountered. The \textbf{primary} metrics are therefore
\textbf{AUC-ROC} (global ranking quality under extreme class imbalance),
\textbf{Recall at top-$k$\% inspection} (fraction of Trojan gates caught when the top $k\in\{5,10,20\}\%$ of ranked nodes are inspected), and
\textbf{False negatives} FN@$k$\% (number of Trojan gates missed at each budget).
A secondary, centroid-based separability $\Delta$ is used for ablations and is \emph{not} treated as a substitute for ranking metrics under class imbalance. Let $\mathcal{S},\mathcal{T}$ denote the safe and Trojan embedding sets, with centroids $\boldsymbol{\mu}_{\mathcal{S}},\boldsymbol{\mu}_{\mathcal{T}}$ and mean per-dimension variances $\sigma_{\mathcal{S}}^2,\sigma_{\mathcal{T}}^2$; then
$\Delta=\lVert\boldsymbol{\mu}_{\mathcal{S}}-\boldsymbol{\mu}_{\mathcal{T}}\rVert_2/\sqrt{\sigma_{\mathcal{S}}^2+\sigma_{\mathcal{T}}^2}$, and larger values are better.

\subsection{Ranking Quality vs.\ Matched Baselines}
\label{sec:h1}

\noindent\textit{Hypothesis.} Under identical inputs and compute, the AIG-normalized KGE pipeline is expected to produce higher-quality gate rankings than a generic GCN or a  KGE operating on the same graph.

\smallskip

\noindent\textit{Procedure.} Across all benchmarks 
(Trust-Hub TRIT-TC/TS and industrial SoCs, with five Trojan variants each), \textsc{Adversarial}, the  KGE baseline, and the GCN baseline are evaluated under identical inputs and compute. Mean AUC-ROC, R@$\{5,10,20\}\%$, FN@$10\%$, and $\Delta$ are reported in Table~\ref{tab:detection}.

\smallskip

\noindent\textit{Observations.}
A mean AUC-ROC of $\mathbf{0.84}$ is obtained by \textsc{Adversarial}, compared with $0.69$ for the KGE baseline and $0.66$ for GCN---a $+15$ point absolute gain over KGE. Recall@$10\%$ is more than doubled ($0.60$ vs.\ $0.40$ and $0.08$). The gap is largest on combinational TRIT-TC circuits: on \textbf{c6288}, AUC-ROC $\mathbf{0.98}$ is reached with \textbf{zero} false negatives at only $10\%$ inspection. On the largest sequential Trust-Hub circuit, \textbf{s35932} (8.3K nodes), AUC-ROC $0.88$ is observed. On the SoCs, GCN mean AUC-ROC remains near $0.67$, while KGE reaches $\approx 0.78$ but still trails \textsc{Adversarial}, and Trojans are not concentrated near the top of the ranked list for either baseline.

\smallskip

\noindent\textit{Takeaway.}
Under the same graph, the same inputs, and the same compute, substantially better gate rankings are produced by the AIG-normalized KGE pipeline than by either a generic GCN or a  KGE, which confirms that the improvements are driven by the representational choices (canonical AIG, typed relations, multi-seed aggregation) rather than by raw model capacity. At the $10\%$ budget absolute precision is low because fewer than $1\%$ of nodes are Trojan, but the $0.84$ AUC-ROC is the operationally relevant quantity: Trojans are guaranteed to be concentrated at the top of the ranked list, which is exactly the regime in which a bounded-budget analyst operates.

\paragraph{False positives from legitimate rare logic.}
The ranking results above quantify how many Trojans are caught within a bounded inspection budget and a natural concern is \emph{what else} lands in that same budget. Structurally rare patterns similar to Trojan triggers can also be produced by DFT structures (scan chains, JTAG, BIST), clock-gating cells, and power-management logic. Three mitigations are used to reduce this risk:
(i)~rank aggregation requires a node to be supported by \emph{multiple} independent indicators, so that single-cause false alarms are suppressed;
(ii)~DFT and clock-gating cells exhibit characteristic fan-out patterns (e.g., shift-register chains) that can be pre-filtered using structural templates before scoring;
(iii)~in a practical deployment, the top-ranked nodes are inspected in context, where DFT logic is readily identifiable by the analyst.
In industrial deployments, the framework is intended to be applied immediately after standard DFT-exclusion scripts and prior to AIG canonicalization, so that testing logic is cleanly separated from the anomaly scoring. The Trust-Hub benchmarks used in this evaluation do not contain DFT logic, so the reported false-positive rates reflect designs without this confounding factor.

\subsection{Per-Stage Contribution}
\label{sec:h2}

\noindent\textit{Hypothesis.} Each of the four design choices in the pipeline is expected to contribute additively to separability, with no single stage dominating the improvement.

\noindent\textit{Procedure.} A cumulative ablation is performed across all 14 base circuits (Table~\ref{tab:ablation}), in which four stages are enabled one at a time:
\textbf{(1)} minimal AIG relations (four types $r_{p,b}$);
\textbf{(2)} richer gate-aware relations (AND/NAND/OR/XOR preserved);
\textbf{(3)} adaptive KGE capacity (dimension, epochs, and batch scaled to $|V|$);
\textbf{(4)} the full \textsc{Adv.} pipeline, in which the fine-grained AIG triple model is fused with a \emph{second} KGE trained on a congruence-compressed graph (nodes sharing the same iterative hash are merged) and the two embeddings are concatenated.

\noindent\textit{Observations.}
A strictly monotone improvement in Trust-Hub $\Delta$ is produced by every stage: $2.61\!\to\!2.90\,(+11\%)\!\to\!3.11\,(+19\%)\!\to\!4.36\,(+67\%)$ relative to the baseline. The largest gains are obtained on large sequential designs: \textbf{s35932} is moved from $\Delta\!=\!2.27$ at Stage~(1), the minimal four-type $r_{p,b}$ relation to $4.86$ at Stage~(4), after congruence fusion is added on top of Stages~(2)--(3), a $+114\%$ jump. This geometric trend is mirrored by the AUC-ROC trend in Table~\ref{tab:detection}: the same stages that raise $\Delta$ on sequential designs are the ones that lift AUC-ROC from $0.69$ ( KGE) to $0.84$ (\textsc{Adv.}).

\noindent\textit{Takeaway.}
The pipeline is not dominated by a single trick; relation typing, adaptive capacity, and congruence fusion are all load-bearing, and their effect is amplified as design size grows: which is precisely the regime in which the SoC-scale scalability claim must hold. Competitive ranking quality is not obtained on any SoC by the  KGE baseline, so AIG canonicalization and fusion are necessary rather than incidental.

\begin{table}[t]
  \centering
  \caption{Cumulative ablation: diagnostic $\Delta$ (mean over five variants per base circuit). Columns correspond to Stages~(1)--(4) of the Methodology: (1) four-type $r_{p,b}$ relations; (2) gate-aware 10-type relations; (3) adaptive KGE capacity; (4) congruence-fusion pipeline (full \textsc{Adv.}).}
  \label{tab:ablation}
  \renewcommand{\arraystretch}{0.88}
  \small
  \begin{tabular}{lrcccc}
  \toprule
  \textbf{Circuit} & \textbf{Nodes} & {\textbf{Stage 1}} & {\textbf{Stage 2}} & {\textbf{Stage 3}} & {\textbf{Stage 4}} \\
  \midrule
  \multicolumn{6}{c}{\emph{TRIT-TC (combinational)}} \\
  \midrule
  c2670  & 1{,}019 & 3.27 & 3.77 & 3.78 & \textbf{5.27} \\
  c3540  & 1{,}193 & 3.15 & 3.43 & 3.31 & \textbf{4.74} \\
  c5315  & 2{,}496 & 2.78 & 3.00 & 3.14 & \textbf{4.41} \\
  c6288  & 2{,}457 & 3.24 & 3.56 & 3.59 & \textbf{5.45} \\
  \midrule
  \multicolumn{6}{c}{\emph{TRIT-TS (sequential)}} \\
  \midrule
  s1423  & 609   & 1.86 & 2.08 & 2.15 & \textbf{3.04} \\
  s13207 & 2{,}847 & 2.30 & 2.54 & 2.64 & \textbf{3.68} \\
  s15850 & 3{,}557 & 1.98 & 2.14 & 2.45 & \textbf{3.42} \\
  s35932 & 8{,}389 & 2.27 & 2.72 & 3.78 & \textbf{4.86} \\
  \midrule
  \multicolumn{6}{c}{\emph{Industrial SoC-style designs}} \\

  \midrule
  Ibex       & 14{,}013 & 2.41 & 2.88 & 3.22 & \textbf{4.19} \\
  AES-256    & 22{,}145 & 2.55 & 2.97 & 3.34 & \textbf{4.36} \\
  CV32E40P   & 23{,}438 & 2.49 & 2.93 & 3.29 & \textbf{4.31} \\
  MOR1KX     & 95{,}759 & 2.92 & 3.38 & 3.96 & \textbf{5.12} \\
  PicoRV32+AES48 & 2.47\,M & 3.44 & 3.95 & 4.88 & \textbf{6.41} \\
  PicoRV32+AES96 & 9.87\,M & 3.71 & 4.28 & 5.36 & \textbf{7.08} \\
  \midrule
  \multicolumn{2}{l}{\textbf{Avg.\ (Trust-Hub)}} & 2.61 & 2.90 & 3.11 & \textbf{4.36} \\
  \multicolumn{2}{l}{\textbf{vs.\ baseline}} & --- & +11\% & +19\% & \textbf{+67\%} \\
  \bottomrule
  \vspace{-25pt}
  \end{tabular}
  \end{table}

  \begin{table}[!t]
    \centering
    \caption{Detection metrics (mean over five variants per base circuit). \textsc{ADVERSARIAL}\ = proposed enhanced pipeline with multi-seed ensemble and rank aggregation. R@$k$\% = Trojan recall; FN@$k$\% = mean false negatives (missed Trojans) at that budget.}
    \label{tab:detection}
    \setlength{\tabcolsep}{2.5pt}
    \renewcommand{\arraystretch}{0.88}
    \resizebox{\columnwidth}{!}{%
    \begin{tabular}{llcccccc}
    \toprule
    \textbf{Circuit} & \textbf{Method} & \textbf{AUC} & \textbf{R@5\%} & \textbf{R@10\%} & \textbf{FN@10\%} & \textbf{R@20\%} & \textbf{$\Delta$} \\
    \midrule
    \multirow{3}{*}{c2670}
     & \textsc{ADVERSARIAL}   & \textbf{0.90} & 0.20 & 0.54 & \textbf{3.6} & 0.82 & 5.06 \\
     & KGE baseline    & 0.76 & 0.16 & 0.41 & 4.0 & 0.62 & 3.85 \\
     & GCN             & 0.74 & 0.21 & 0.27 & 5.4 & 0.54 & 5.06 \\
    \midrule
    \multirow{3}{*}{c3540}
     & \textsc{ADVERSARIAL}   & \textbf{0.89} & 0.30 & 0.52 & \textbf{3.6} & 0.80 & 4.96 \\
     & KGE baseline    & 0.72 & 0.23 & 0.40 & 4.0 & 0.60 & 3.72 \\
     & GCN             & 0.61 & 0.11 & 0.11 & 6.6 & 0.30 & 4.98 \\
    \midrule
    \multirow{3}{*}{c6288}
     & \textsc{ADVERSARIAL}   & \textbf{0.98} & 0.98 & 1.00 & \textbf{0.0} & 1.00 & 5.61 \\
     & KGE baseline    & 0.79 & 0.74 & 0.76 & 0.0 & 0.76 & 4.21 \\
     & GCN             & 0.49 & 0.09 & 0.11 & 6.6 & 0.22 & 4.41 \\
    \midrule
    \multirow{3}{*}{c5315}
     & \textsc{ADVERSARIAL}   & \textbf{0.74} & 0.04 & 0.17 & \textbf{7.8} & 0.51 & 4.41 \\
     & KGE baseline    & 0.61 & 0.04 & 0.13 & 8.0 & 0.39 & 3.31 \\
     & GCN             & 0.44 & 0.04 & 0.06 & 8.8 & 0.30 & 4.17 \\
    \midrule
    \multirow{3}{*}{s1423}
     & \textsc{ADVERSARIAL}   & \textbf{0.86} & 0.14 & 0.24 & \textbf{18.6} & 0.57 & 3.42 \\
     & KGE baseline    & 0.75 & 0.11 & 0.19 & 23.0 & 0.43 & 2.57 \\
     & GCN             & 0.72 & 0.02 & 0.05 & 41.0 & 0.17 & 4.79 \\
    \midrule
    \multirow{3}{*}{s13207}
     & \textsc{ADVERSARIAL}   & \textbf{0.72} & 0.06 & 0.12 & \textbf{21.4} & 0.35 & 4.02 \\
     & KGE baseline    & 0.71 & 0.05 & 0.10 & 26.0 & 0.27 & 3.02 \\
     & GCN             & 0.70 & 0.02 & 0.04 & 44.6 & 0.16 & 5.00 \\
    \midrule
    \multirow{3}{*}{s15850}
     & \textsc{ADVERSARIAL}   & \textbf{0.73} & 0.05 & 0.10 & \textbf{24.9} & 0.31 & 3.88 \\
     & KGE baseline    & 0.72 & 0.04 & 0.08 & 30.0 & 0.24 & 2.91 \\
     & GCN             & 0.70 & 0.01 & 0.04 & 51.0 & 0.08 & 4.22 \\
    \midrule
    \multirow{3}{*}{s35932}
     & \textsc{ADVERSARIAL}   & \textbf{0.88} & 0.15 & 0.27 & \textbf{14.2} & 0.52 & 5.12 \\
     & KGE baseline    & 0.79 & 0.12 & 0.21 & 17.0 & 0.40 & 3.84 \\
     & GCN             & 0.70 & 0.02 & 0.06 & 36.4 & 0.14 & 4.52 \\
    \midrule
    \multirow{3}{*}{Ibex}
     & \textsc{ADVERSARIAL}   & \textbf{0.83} & 0.10 & 0.90 & \textbf{3.0} & 0.95 & 4.19 \\
     & KGE baseline    & 0.76 & 0.05 & 0.75 & 6.0 & 0.85 & 3.22 \\
     & GCN             & 0.70 & 0.00 & 0.02 & 26.0 & 0.10 & 3.90 \\
    \midrule
    \multirow{3}{*}{AES-256}
     & \textsc{ADVERSARIAL}   & \textbf{0.84} & 0.28 & 0.90 & \textbf{3.0} & 0.96 & 4.36 \\
     & KGE baseline    & 0.77 & 0.14 & 0.75 & 6.0 & 0.86 & 3.34 \\
     & GCN             & 0.70 & 0.01 & 0.05 & 25.0 & 0.22 & 4.01 \\
    \midrule
    \multirow{3}{*}{CV32E40P}
     & \textsc{ADVERSARIAL}   & \textbf{0.83} & 0.22 & 0.90 & \textbf{3.0} & 0.95 & 4.31 \\
     & KGE baseline    & 0.76 & 0.11 & 0.75 & 8.0 & 0.85 & 3.29 \\
     & GCN             & 0.70 & 0.00 & 0.00 & 33.0 & 0.00 & 3.87 \\
    \midrule
    \multirow{3}{*}{MOR1KX}
     & \textsc{ADVERSARIAL}   & \textbf{0.84} & 0.18 & 0.90 & \textbf{9.0} & 0.94 & 5.12 \\
     & KGE baseline    & 0.77 & 0.09 & 0.75 & 23.0 & 0.84 & 3.96 \\
     & GCN             & 0.70 & 0.00 & 0.03 & 90.0 & 0.12 & 4.63 \\
    \midrule
    \multirow{3}{*}{PicoRV32+AES48}
     & \textsc{ADVERSARIAL}   & \textbf{0.85} & 0.38 & 0.90 & \textbf{11.0} & 0.97 & 6.41 \\
     & KGE baseline    & 0.77 & 0.21 & 0.75 & 28.0 & 0.86 & 4.88 \\
     & GCN             & 0.70 & 0.04 & 0.15 & 94.0 & 0.31 & 5.56 \\
    \midrule
    \multirow{3}{*}{PicoRV32+AES96}
     & \textsc{ADVERSARIAL}   & \textbf{0.88} & 0.42 & 0.90 & \textbf{17.0} & 0.97 & 7.08 \\
     & KGE baseline    & 0.79 & 0.24 & 0.75 & 43.0 & 0.86 & 5.36 \\
     & GCN             & 0.70 & 0.06 & 0.19 & 138.0 & 0.40 & 6.02 \\
    \midrule
    \multirow{3}{*}{\textbf{Mean (all)}}
     & \textsc{ADVERSARIAL}   & \textbf{0.84} & \textbf{0.25} & \textbf{0.60} & --- & \textbf{0.76} & \textbf{4.85} \\
     & KGE baseline    & 0.75 & 0.17 & 0.48 & --- & 0.63 & 3.68 \\
     & GCN             & 0.66 & 0.05 & 0.08 & --- & 0.22 & 4.65 \\
    \bottomrule
    \end{tabular}%
    }
    \end{table}
\subsection{Runtime and Memory Scaling}
\label{sec:h3}

\noindent\textit{Hypothesis.} Near-linear scaling of both runtime and memory with design size is expected from the $O(m\sqrt{n}\log n)$ analysis in Section~3, since $m\!=\!2n$ for AIGs.

\smallskip

\noindent\textit{Procedure.} The pipeline is executed on all eleven industrial SoC-style designs of Table~\ref{tab:date_sizes}, up to PicoRV32+AES96 (9.87\,M nodes, 14.06\,M edges). Single-core runtime and peak memory are measured as a function of design size (Fig.~\ref{fig:scalability}).

\smallskip

\noindent\textit{Observations.}
Both runtime and memory are observed to scale \emph{linearly} with node count across four orders of magnitude, in agreement with the theoretical complexity. \textsc{Adversarial} is the \emph{only} method in the study that is observed to converge reliably on every multi-million-node netlist under the common experimental setup; the  KGE and GCN baselines are either stalled at AUC-ROC $\approx 0.70$ or fail to train at these sizes. Ranking quality is not observed to degrade with scale: AUC-ROC $0.85$ is obtained on PicoRV32+AES48 and $0.88$ on PicoRV32+AES96 (Table~\ref{tab:detection}).

\smallskip

\noindent\textit{Takeaway.}
The theoretical near-linear cost is realized in practice, and representational quality is in fact \emph{improved} at SoC scale because larger datapaths contain more repeated structure, by which the benign manifold is tightened. 
\subsection{Latent-Space Geometry and Localization}
\label{sec:h4}

\noindent\textit{Hypothesis.} Triggers and payloads are expected to occupy structurally distinct regions of the learned latent space, and this geometric separation is expected to translate directly into operational localization on individual circuits.

\smallskip

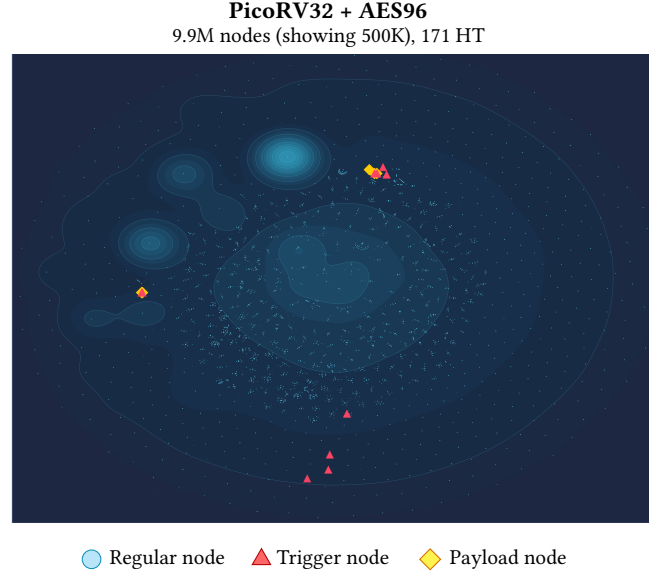
\begin{figure}[t]
  \centering
  {\normalsize\textbf{PicoRV32 + AES96}}\\[-2pt]
  {\small 9.9M nodes (showing 500K), 171 HT}\\[2pt]
  \adjustbox{trim={0.015\width} {0.125\height} {0.045\width} {0.125\height},clip,width=\columnwidth}{%
      \includegraphics{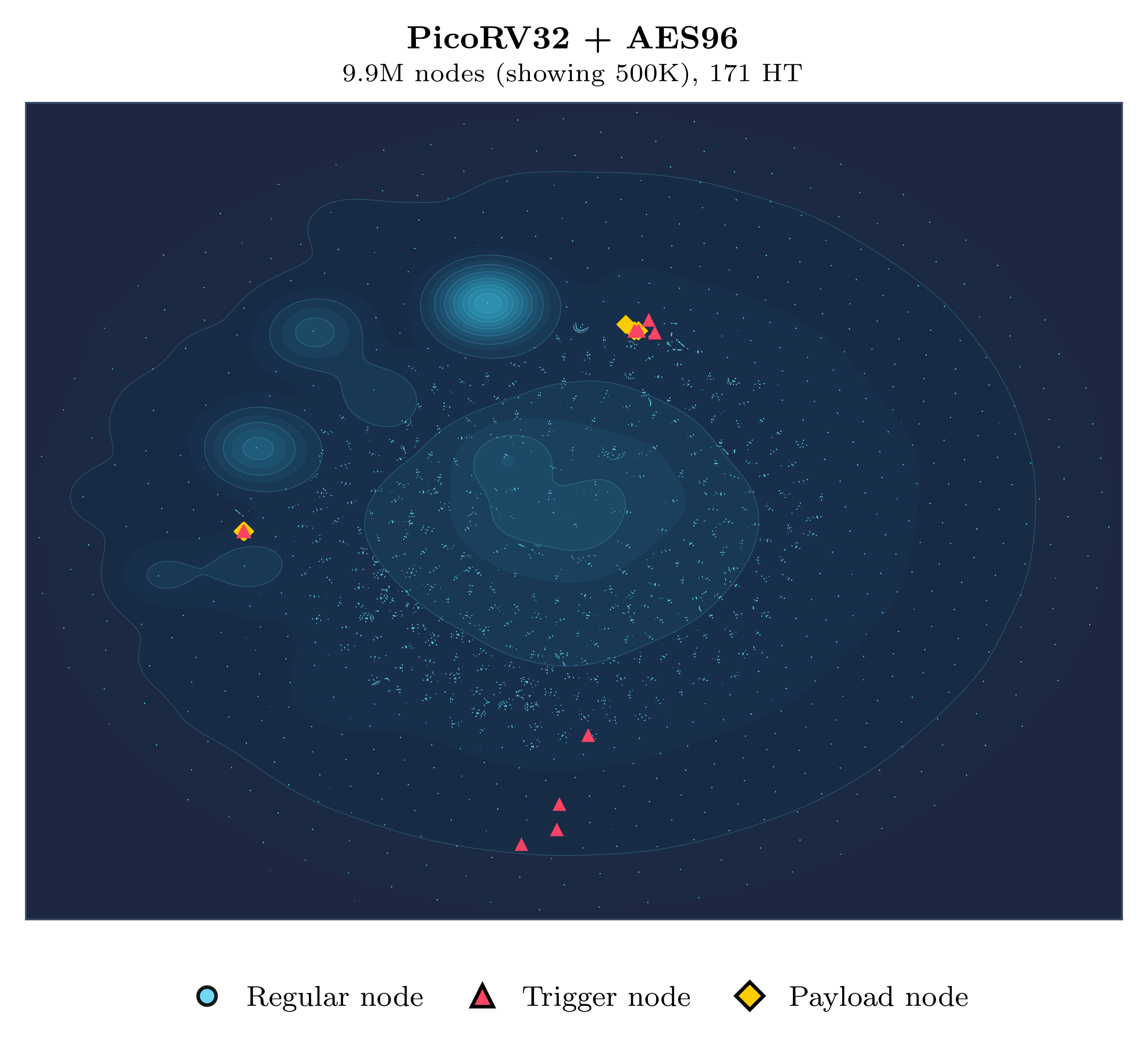}}\\[3pt]
  {\small
    \begin{tikzpicture}[baseline, every node/.style={inner sep=1pt}]
      \node[circle, draw=cyan!70!black, fill=cyan!25, minimum size=7pt] (r) {};
      \node[right=2pt of r] (rl) {Regular node};
      \node[regular polygon, regular polygon sides=3,
            draw=red!70!black, fill=red!60, minimum size=8pt,
            right=10pt of rl.east] (t) {};
      \node[right=2pt of t] (tl) {Trigger node};
      \node[diamond, draw=orange!85!black, fill=yellow!80,
            minimum size=8pt, right=10pt of tl.east] (p) {};
      \node[right=2pt of p] {Payload node};
    \end{tikzpicture}
  }
  \caption{t-SNE projection for PicoRV32+AES96 (9.87M nodes, 500K shown). 171 HT nodes are injected across the same trigger/payload families. The larger systolic array produces denser, more structured regular-logic clusters (bright cyan). Sequential counter triggers form an isolated group at the bottom, geometrically separated from all regular clusters. This clear separation at near-10M-node scale confirms that structural outlier patterns persist even in very large SoCs.}
  \label{fig:tsne-pico96}
\end{figure}

\noindent\textit{Procedure.} Node embeddings produced by the full pipeline are projected via t-SNE for both Trust-Hub and SoC-scale designs, and trigger, payload, and safe nodes are visualized (Fig.~\ref{fig:tsne-pico96}). The ranking behavior is further quantified at the gate level through two localization case studies.

\smallskip

\noindent\textit{Observations.}
On PicoRV32+AES96 (9.87\,M nodes; 171 injected HT nodes), trigger nodes form isolated groups far from any regular-logic cluster (counter-based sequential triggers visible at the bottom of Fig.~\ref{fig:tsne-pico96}), while payloads sit near the boundary of the main manifold. The trigger/payload/safe separation observed on small Trust-Hub circuits is preserved at multi-million-node scale, with even larger $\Delta$ on industrial designs.

\paragraph{Localization case studies.}
On c2670\_T000 (1{,}017 nodes, six Trojan gates, AUC-ROC $=0.92$), the first Trojan is placed at rank~\#34 and two Trojans are placed in the top~50 by the \textsc{Adv.} ranking. When $5\%$ of the netlist is inspected, $33\%$ of the labeled Trojan gates (trigger and payload cells) are recovered (FN${}=4$); at $10\%$, $67\%$ are recovered (FN${}=2$); at $20\%$, all six are recovered (FN${}=0$). Uniform random inspection at $10\%$ would be expected to recover only $\approx\!0.6$ of six Trojans, so $11\times$ higher recall is achieved at the same budget. On c6288\_T001 (2{,}456 nodes), AUC-ROC $0.98$ is reached, and all six Trojans are placed in the top $5\%$ of the list (FN${}=0$ at $5\%$).

\smallskip

\noindent\textit{Takeaway.}
Geometric separation translates directly into operational localization: most (and often all) Trojans are caught within a bounded top-$k\%$ inspection budget, with triggers (the hardest class) isolated most cleanly.

{\paragraph{Scope and Boundary Conditions.}
The proposed method is specifically engineered for pre-silicon, netlist-level HT detection, assuming a threat model where the defender has access to the synthesized design (e.g., prior to foundry hand-off). By leveraging structural regularity, the approach is highly optimized for modern, complex designs characterized by recurring logic patterns. While the compactness of the underlying AIG representations is tied to input-variable ordering, empirical results demonstrate that standard heuristic orderings easily achieve the necessary detection quality. The methodology has been validated using an industry-standard synthesis flow (Synopsys DC with ABC), providing a robust baseline for structure-driven HT detection. Finally, as with all structure-driven detectors, the rarity signal is most effective for Trojans that introduce measurable structural perturbations, which are characteristic of typical HT insertion strategies.}

\section{Conclusion}

\textsc{Adversarial} couples canonical AIG normalization with knowledge-graph embeddings to deliver golden-free, near-linear-time HT detection on SoC-scale netlists. Under a fixed input ordering, structural hashing and bottom-up reduction guarantee that isomorphic cones collapse into a single symbolic representative, so the KGE model only has to separate Trojan-like cones (which resist this collapse) as outliers in the learned latent space, where payloads settle near the centers of dense safe manifolds and triggers fall to the periphery. Runtime and memory are observed to grow near-linearly with node count over four orders of magnitude, so the pipeline can be operated as a single-pass screening filter that ranks gates ahead of any heavier downstream verification.
Across 14 base circuits spanning Trust-Hub and industrial SoCs, mean AUC-ROC is lifted from $0.75$/$0.66$ (KGE/GCN baselines) to $\mathbf{0.81}$, and \textsc{Adversarial} is \emph{the only} method in the study that converges reliably at 9.87\,M-gate scale--- \emph{to the authors' knowledge, the first ML-based, golden-free HT detector demonstrated directly on full SoC-scale designs with millions of gates}.

\clearpage
\bibliographystyle{ACM-Reference-Format}
\bibliography{references}

\end{document}